\documentclass[10pt,journal]{IEEEtran}
\usepackage[pdftex]{graphicx}
\DeclareGraphicsExtensions{.pdf,.jpg,.png}
\usepackage{cite}
\usepackage[cmex10]{amsmath}
\usepackage{amssymb}
\usepackage{graphicx}
\usepackage{cases}
\usepackage{mdwmath}
\usepackage{mdwtab}
\usepackage{array}
\usepackage{url}
\usepackage[ruled,vlined]{algorithm2e}
\usepackage{booktabs}
\usepackage{threeparttable}
\usepackage{multirow}
\usepackage{color}
\usepackage{caption}

\def\ie{{\em i.e.}}
\def\eg{{\em e.g.}}
\def\etal{{\em et al.}}

\newcommand{\myPara}[1]{\vspace{.05in}\noindent\textbf{#1}}

\newcommand{\bl}[1]{\textbf{#1}}
\newcommand{\ul}[1]{\underline{#1}}

\newcommand{\mc}[1]{\mathcal{#1}}
\newcommand{\mb}[1]{\mathbb{#1}}

\newcommand{\bm}[1]{\mbox{\boldmath{$#1$}}}

\usepackage{color}
\begin{document}

\title{Low-resolution Face Recognition in the Wild \\ via Selective Knowledge Distillation}
\author{Shiming Ge,~\IEEEmembership{Senior Member,~IEEE,}
    Shengwei Zhao, Chenyu Li and
    Jia~Li,~\IEEEmembership{Senior Member,~IEEE}
\thanks{S. Ge is with the Institute of Information Engineering, Chinese Academy of Sciences, Beijing, 100095, China. E-mail: geshiming@iie.ac.cn}
\thanks{S. Zhao and C. Li are with the Institute of Information Engineering, Chinese Academy of Sciences, and School of Cyber Security at University of Chinese Academy of Sciences.}
\thanks{J.~Li is with State Key Laboratory of Virtual Reality Technology and Systems, School of Computer Science and Engineering, Beihang University. He is also with Beijing Advanced Innovation Center for Big Data and Brain Computing, Beihang University, Beijing, 100191, China. E-mail:jiali@buaa.edu.cn }
\thanks{J. Li is the corresponding author.}
}

\markboth{~}%
{Ge \MakeLowercase{\textit{et al.}}: Bare Demo of IEEEtran.cls for Journals}

\maketitle

\begin{abstract}
  Typically, the deployment of face recognition models in the wild needs to identify low-resolution faces with extremely low computational cost. To address this problem, a feasible solution is compressing a complex face model to achieve higher speed and lower memory at the cost of minimal performance drop. Inspired by that, this paper proposes a learning approach to recognize low-resolution faces via selective knowledge distillation. In this approach, a two-stream convolutional neural network (CNN) is first initialized to recognize high-resolution faces and resolution-degraded faces with a teacher stream and a student stream, respectively. The teacher stream is represented by a complex CNN for high-accuracy recognition, and the student stream is represented by a much simpler CNN for low-complexity recognition. To avoid significant performance drop at the student stream, we then selectively distil the most informative facial features from the teacher stream by solving a sparse graph optimization problem, which are then used to regularize the fine-tuning process of the student stream. In this way, the student stream is actually trained by simultaneously handling two tasks with limited computational resources: approximating the most informative facial cues via feature regression, and recovering the missing facial cues via low-resolution face classification. Experimental results show that the student stream performs impressively in recognizing low-resolution faces and costs only $0.15$MB memory and runs at $418$ faces per second on CPU and $9,433$ faces per second on GPU.
\end{abstract}



\IEEEpeerreviewmaketitle

\section{Introduction}

\begin{figure}[t]
\begin{center}
   \includegraphics[width=1.0\linewidth]{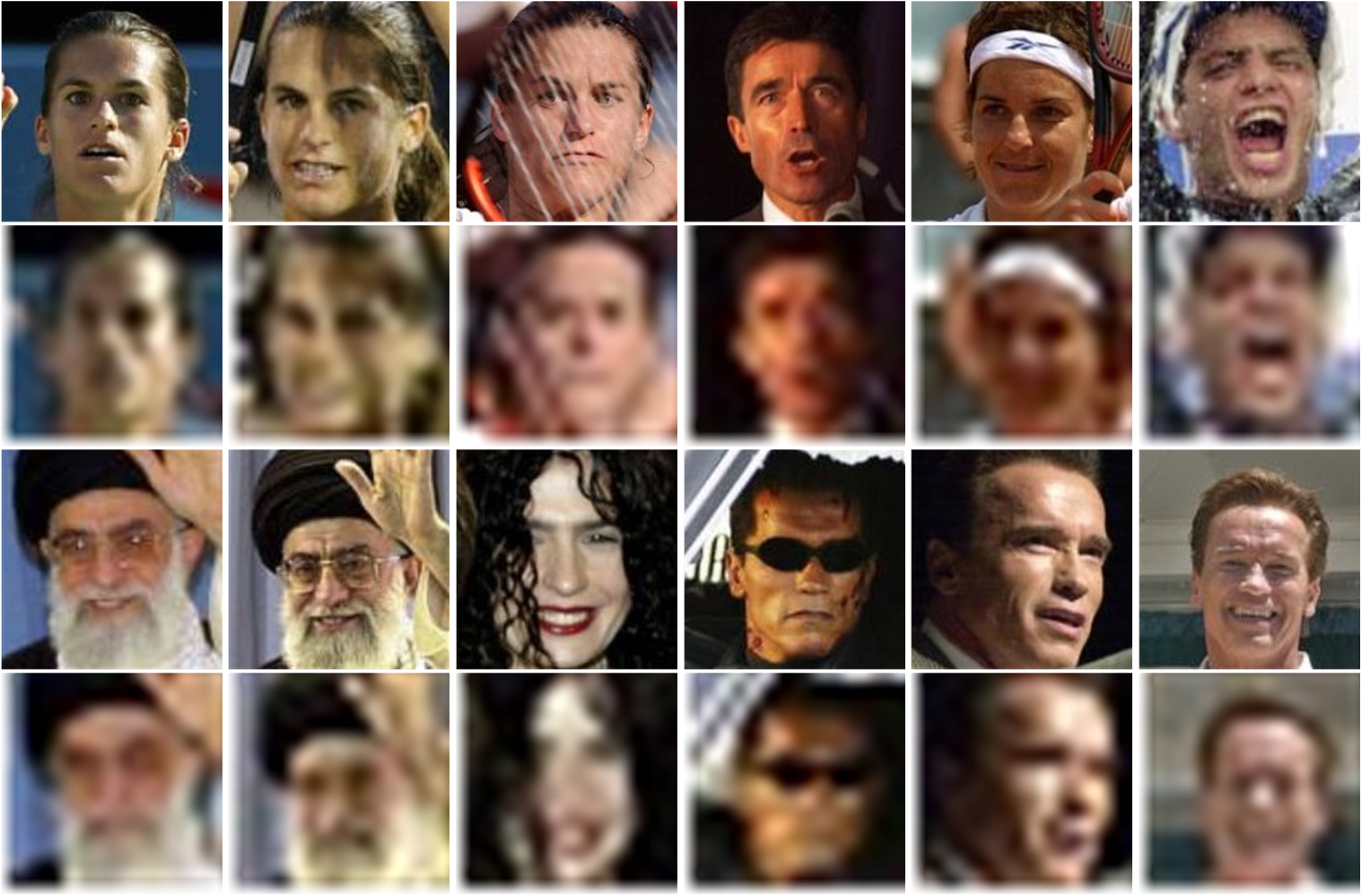}
\end{center}
   \caption{Low-resolution faces can become recognizable for subjects that are familiar with the corresponding high-resolution faces. Although low-resolution faces may have low quality, blurry textures, poor illumination and diversified occlusions, the knowledge from high-resolution faces can help to extract discriminative features for efficient face recognition.}
\label{Fig:example}
\end{figure}

Face, a fundamental attribute that distinguishes one subject from another, needs to be recognized many times everyday in modern computer vision and multimedia applications. Among these applications, many well-known face recognition models~\cite{Taigman2014CVPR,Sun2014CVPR,Schroff2015CVPR,Parkhi2015BMVC} need to be re-deployed on mobile phones~\cite{amos2016openface} or even smart cameras~\cite{pentland2000face} to meet the real-world requirements that aim to identify low-resolution faces with extremely low computational cost and memory footprint (\ie, face recognition in the wild~\cite{liu2017enhance}). Toward this end, it is necessary to explore a feasible solution that can address a key challenge in face recognition: how to convert an existing complex face model into a more efficient one that still works well on low-resolution faces without remarkable loss of recognition accuracy?

Compared with high-resolution faces, low-resolution faces have their unique visual attributes. As shown in Fig.~\ref{Fig:example}, many details are missing in low-resolution faces. However, they are still recognizable for subjects who are familiar with the corresponding high-resolution faces, implying that the neural systems of human beings may have the capability of recovering missing details of familiar faces. Inspired by this fact, many existing low-resolution face models have been proposed, which can be roughly grouped into two categories: the hallucination category and the embedding category.

\begin{figure*}[t]
\begin{center}
   \includegraphics[width=1.0\linewidth]{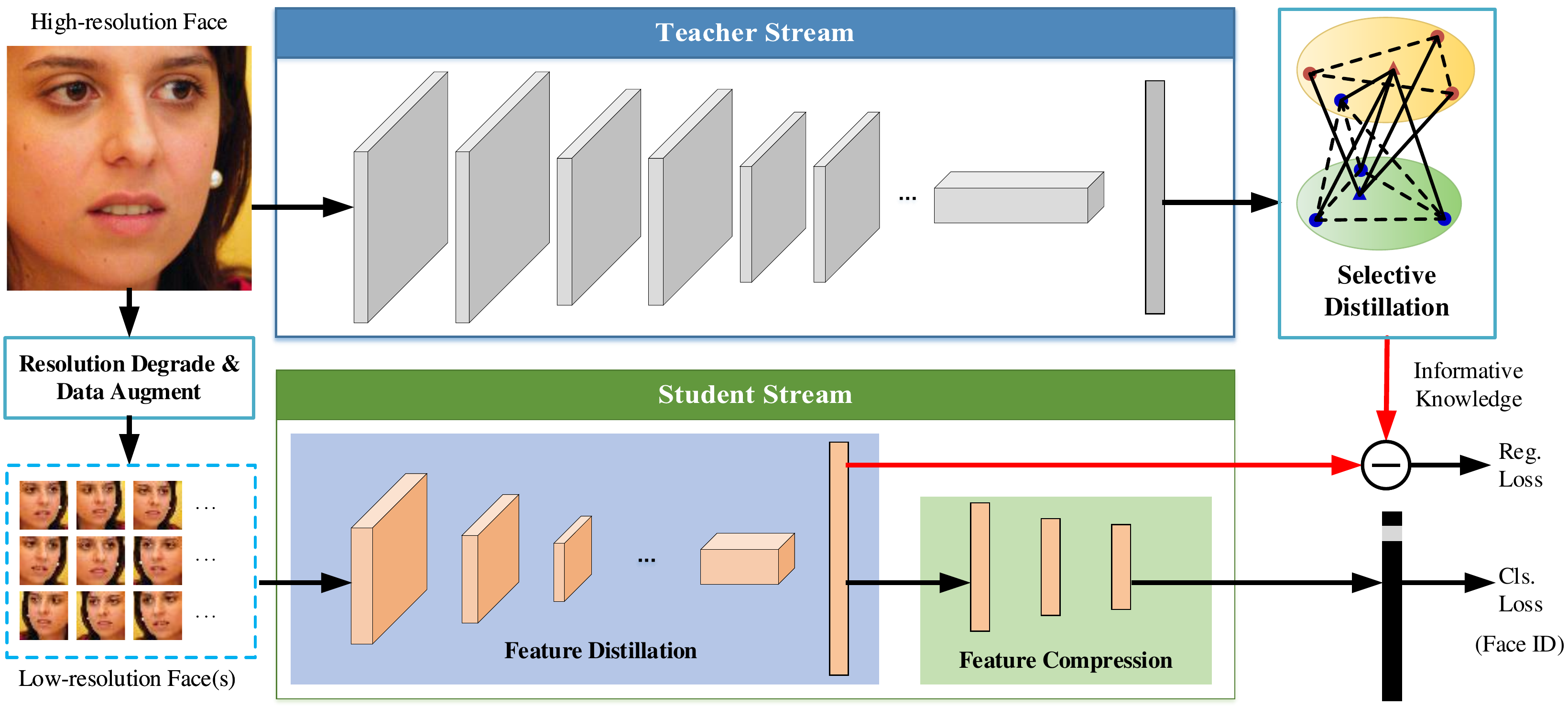}
\end{center}
   \caption{The framework of our approach. The distillation process consists of three stages. In the first stage, we initialize the teacher and student streams to recognize high-resolution faces and their low-resolution versions, respectively. In the second stage, we selectively distil only the most informative facial features from the teacher stream by solving a sparse graph optimization problem. In the third stage, the selected features are used to regularize the fine-tuning process of the student stream. In this manner, the student stream for recognizing low-resolution faces in the wild is actually trained by simultaneously handling two tasks with limited computational resources: selective feature approximation and low-resolution face identification.}
\label{fig:framework}
\end{figure*}

Models in the hallucination category propose reconstructing the high-resolution faces before recognition~\cite{kolouri2015transport,jian2015simultaneous,yang2015recognition}. For example,
Kolouri~\etal~\cite{kolouri2015transport} described a transport-based single frame super-resolution method to automatically construct a nonlinear Lagrangian model of high-resolution facial appearance. After that, the low-resolution facial image was enhanced by finding the model parameters that best fit the given low-resolution data.
Jian~\etal~\cite{jian2015simultaneous} observed that the singular values of a face image at different resolutions have approximately linear relationship. Based on this observation, they first applied singular value decomposition for face representation to learn the mapping function between low-resolution and high-resolution face pairs, and then performed both hallucination and recognition of low-resolution faces simultaneously. Similar method proposed by Yang~\etal~\cite{yang2015recognition} used sparse representation to perform joint hallucination and recognition, which can synthesize person-specific versions of low-resolution faces with recognition guarantee. Typically, these approaches exhibit impressive performance in recognizing the reconstructed high-resolution faces, while the super-resolution operation often brings in additional computational cost and leads to low recognition speed.

Different from the hallucination-based models, models in the embedding category directly extract discriminative features from low-resolution faces by using various external face contexts. For example, Biswas~\etal~\cite{biswas2012multidimensional} proposed embedding low-resolution facial images into an Euclidean space such that the distances between them in the transformed space can well approximate the best distances of high-resolution faces. Ren~\etal~\cite{ren2012coupled} proposed coupled kernel embedding to map the facial images with different resolutions onto an infinite subspace. The recognition process was then carried out in the new space by minimizing the dissimilarities captured by their kernel gram matrices in the low-resolution and high-resolution spaces. Generally speaking, the most important process in the embedding-based approaches is transferring the knowledge from high-resolution faces to low-resolution ones. However, a key issue that needs to be carefully addressed in this process is correctly transferring only the desired knowledge rather than all of them from high-resolution domain to low-resolution domain. Such selective knowledge transfer is one of the most important challenges in converting existing face models into more efficient ones that also work well on low-resolution faces.

Inspired by this fact, we propose a selective knowledge distillation approach for low-resolution face recognition in the wild. As shown in Fig.~\ref{fig:framework}, a two-stream CNN is first trained to simultaneously recognize high-resolution faces and their resolution-degraded versions by using two streams. The two streams consist of a teacher stream that operates on high-resolution faces, and a student stream that is much simpler for low-resolution face recognition. To ensure that the student stream has comparable recognition performance with the teacher stream, we then selectively distil only the most informative facial features from the teacher stream by solving a sparse graph optimization problem, which are then used to regularize the fine-tuning process of the student stream. In this way, the student stream is actually trained by simultaneously handling two tasks with limited computational resources: approximating the most informative facial cues via feature regression, and recovering the missing facial cues via low-resolution face classification. Note that the teacher stream can adopt any architecture of existing deep face models, implying that the proposed approach can convert any existing face model into a much simpler one with higher speed and lower memory at the cost of minimal performance drop. Experimental results on four public datasets show that the student stream performs impressively in recognizing faces at extremely low resolutions. In particular, it uses only $0.15$MB memory and runs at about $418$ faces per second on a single CPU thread or $9,433$ faces per second on GPU.

The main contributions of this paper are summarized as follows.
1)~We propose a face model compression method via selective knowledge distillation, which can greatly reduce model size and complexity without remarkable performance drop; 2)~We propose graph-based optimization algorithm that can extract the most discriminative facial features from existing face models, which can be used to supervise the training process of low-resolution face models; and 3)~We conduct comprehensive experiments to show that the compressed model can achieve an extremely high recognition speed with a comparable accuracy with the state-of-the-art high-resolution face models.

The rest of this paper is organized as follows: Section~\uppercase\expandafter{\romannumeral2} reviews related works and Section~\uppercase\expandafter{\romannumeral3} presents the selective knowledge distillation approach. Extensive experiments are conducted in Section~\uppercase\expandafter{\romannumeral4} to evaluate the proposed approach, and the paper is concluded in Section~\uppercase\expandafter{\romannumeral5}.

\section{Related Works}
The approach we proposed in this paper aims to distil knowledge from complex face models for low-resolution face recognition. Therefore, we briefly review related works from three aspects, including the general face recognition models, low-resolution face recognition and knowledge distillation.

\subsection{General Face Recognition Models}
Recently, the general face recognition technique has evolved from the classic shallow frameworks \cite{ahonen2004face,Prince2007Probabilistic} to deep ones~\cite{Taigman2014CVPR,Wen2016ECCV,Schroff2015CVPR,Parkhi2015BMVC,Tran2017CVPR,Zhang2017RangeLoss,Hu2017ICCV} with impressive performance improvements. For the deep approaches, a key factor to distinguish them is the loss functions they adopted. For example, DeepFace~\cite{Taigman2014CVPR} is an early attempt to ensemble Convolutional Neural Networks (CNNs) by building 3D faces with identification loss. After that, various loss functions have been proposed for training face recognition CNNs, such as triplet loss~\cite{Schroff2015CVPR,Parkhi2015BMVC}, center loss~\cite{Wen2016ECCV} and range loss~\cite{Zhang2017RangeLoss}. In \cite{Hu2017ICCV}, the tasks of identifying faces and their attributes were simultaneously considered to enhance the recognition performance. For the DeepID series, several small CNNs using different facial patches were first separately trained in \cite{Sun2014CVPR}, and its subsequent works incorporate face verification signals \cite{Sun2014NIPS} and change the base networks~\cite{sun2015deepid3} to increase accuracy.

Generally speaking, these deep models have achieved impressive performance in recognizing general faces. As shown in Tab.~\ref{Table:models}, however, many of such generic models have a large amount of parameters, high dimensional feature representations and complex classification function for inference. The complexity of these models prevent them from being directly deployed in the wild where the computational resource is limited. Although DeepID series take low-resolution faces as the input, the unique attributes of low-resolution faces are not explored. To further enhance the performance of low-resolution face recognition, it is necessary to explore the missing features during the resolution degradation.

\begin{table}[t]\small
\centering
\caption{Representative deep models for general face recognition. \#Train: number of training images, Res.: input face resolution, Dim.: output feature dimension, \#Lyr: number of network layers, \#Par: number of model parameters}
\begin{tabular}{rcrrrr}
\hline
Model                                 &\#Train &Res.           &Dim.     &\#Lyr    &\#Par\\
\hline
DeepFace~ ~\cite{Taigman2014CVPR}     &4.4M    &152$\times$152 &4,096   &8        &120M\\
DeepID~ ~\cite{Sun2014CVPR}            &203K    &39$\times$31   &9,600   &7        &17M\\
DeepID2~\cite{Sun2014NIPS}             &203K    &55$\times$47   &4,500   &7        &10M\\
FaceNet~ ~\cite{Schroff2015CVPR}      &260M    &96$\times$96   &128     &22       &140M\\
VGGFace~ ~\cite{Parkhi2015BMVC}       &2.6M    &224$\times$224 &4,096   &16       &138M\\
CenterLoss\cite{Wen2016ECCV}          &700K    &112$\times$96  &1,024   &7        &6M\\
CCN-3DMM\cite{Tran2017CVPR}           &500K    &100$\times$100 &198     &101      &30M\\
GTNN~\cite{Hu2017ICCV}                &6.0M    &128$\times$128 &512     &10       &3M\\
RangeLoss\cite{Zhang2017RangeLoss}    &1.5M    &112$\times$96  &1,024   &7        &6M\\
VGGFace2\cite{cao2018fg}              &3.31M   &224$\times$224 &2,048   &50       &21.7M\\
\hline
\end{tabular}
\label{Table:models}
\end{table}

\subsection{Low-Resolution Face Recognition}

Typically, there are two ways for low-resolution face recognition. The hallucination category aims to reconstruct high-resolution faces before recognition, while the embedding category proposes extracting features directly from low-resolution faces via the embedding schema. In the hallucination category, Kolouri~\etal~\cite{kolouri2015transport} constructed a nonlinear Lagrangian model of high-resolution facial appearance and then found the model parameters that best fit the low-resolution faces. Jian~\etal~\cite{jian2015simultaneous} proposed a framework based on singular value decomposition and performed face hallucination and recognition simultaneously. In \cite{yang2015recognition}, a joint face hallucination and recognition framework was proposed based on sparse representation. This framework can synthesize person-specific low-resolution faces for recognition. In \cite{uiboupin2016facial}, a system was proposed to recognize faces by using sparse representation with the specific dictionary involving many natural and facial images. Moreover, deep models like \cite{dong2014learning} and \cite{ledig2016photo} can generate extremely realistic high-resolution images from low-resolution faces. However, the speed of such hallucination or super-resolution based approaches may be a little slow due to the complex high-resolution face reconstruction process, which hinders their direct deployment in real-world scenarios with limited computational resources.

Instead of reconstructing high-resolution faces, a more direct approach is embedding low-resolution faces into various external contexts to recover the missing features during resolution degradation. Inspired by that, some approaches proposed transforming both high-resolution and low-resolution faces into a unified feature space for matching~\cite{zou2012very,zhang2015coupled,jiang2016cdmma,wang2016pose,xing2016couple,shi2015local,haghighat2017low}, while in \cite{li2015multi,pong2014multi} the multi-scale (multi-resolution) faces were simultaneously analyzed to extract better features. In~\cite{mudunuri2016low}, the multidimensional scaling was adopted to learn a common transformation matrix to simultaneously transform the facial features of low-resolution and high-resolution training images. Shekhar~\etal~\cite{shekhar2017synthesis} proposed a joint sparse coding technique for robust recognition at low-resolution, while Wang~\etal~\cite{Wang2016CVPR} attempted to solve very low resolution recognition problem using deep learning methods. In \cite{herrmann2016low}, CNNs were adopted with a manifold-based track comparison strategy for low-resolution face recognition in videos.

From these approaches, we find that the core idea of the embedding-based approaches is transferring (or making use of) the knowledge from high-resolution faces. As a result, the performance in low-resolution face recognition is mainly influenced by two key factors: what knowledge to transfer and how to make use of it. In other words, the desired knowledge should be selectively distilled from high-resolution data (or models) and guide the low-resolution face recognition process in a right way. This is also the core idea of this paper.

\subsection{Knowledge Distillation}
Instead of mining the knowledge from high-resolution faces, another way to obtain a low-resolution face model (\ie, the student network) is distilling such knowledge directly from pre-trained complex face models (\ie, the teacher network). With the development of much deeper and wider networks, such distillation technique has been adopted in many works \cite{bucilua2006kdd,Hinton2014NIPSW,Romero2015ICLR,chen2016iclr,Luo2016AAAI,li2016eccv,kim2016sequence,urban2017iclr,chen2017nips,chen2018DarkRank,chen2018WAE,zhou2017Rocket} to compress a complex model (or an ensemble) into a simpler model that is much easier to deploy. Among these works, Luo~\etal~\cite{Luo2016AAAI} utilised the learned knowledge of a large teacher network or the ensemble of some networks as the supervision to train a compact student network for face recognition. In their approach, the most relevant neurons for face recognition were selected at the higher hidden layers for knowledge transfer. Lopez-Paz~\etal~\cite{lopezpaz2016iclr} proposed the general distillation framework to combine distillation and learning with privileged information. Su and Maji~\cite{su2017bmvc} proposed cross quality distillation to learn models for recognizing low-resolution images, non-localized objects and line-drawings by using labeled high-resolution images, labeled localized objects and color images, respectively. Radosavovic~\etal~\cite{radosavovic2018cvpr} proposed data distillation to ensemble predictions from multiple transformations of unlabeled data to automatically generate new training annotations.

To sum up, the core component of knowledge distillation is the trade-off between speed and performance, and such a technique provides an opportunity to convert many complex models into simple models that can be deployed in the wild. Note that in this study we not only try to distil complex face models into simple ones, but also explore the feasibility of using resolution-degraded faces as the input to further speed up the recognition speed while maintaining the recognition accuracy. In this way, the challenges in low-resolution face recognition and knowledge distillation are simultaneously addressed with a single framework.

\section{The Approach}

Our two-stream knowledge distillation framework consists of a teacher stream and a student stream (see Fig.~\ref{fig:framework}). The teacher stream can adopt any complex face recognition neural networks that have been previously trained (and the training data may be no longer available). The distillation process aims to learn a simple and compact student stream that imitates the teacher stream for its practical deployment in real-world scenario.

The learning process consists of three stages: 1)~the Initialization stage initializes the teacher stream by taking a complex CNN or an ensemble of several CNNs pre-trained on high-resolution face images, and the student stream by classifying low-resolution face images with identity labels; 2)~the Selection stage extracts the most informative knowledge from the teacher stream where the ``right'' knowledge is selected while the ``wrong'' one is wiped out; and 3)~the Fine-tuning stage transfers the selective knowledge from teacher and low-resolution face images to co-supervise the fine-tuning progress of the student stream by jointly performing feature regression and face identity classification. More details of the three stages are described as follows.

\subsection{Definition}
For the sake of simplicity, we define the key components of the two-stream CNNs as follows:

\textbf{Teacher stream}. The teacher stream $\phi_t(\mathcal{F};\mathbb{W}_t)$ is a complex CNN (or an ensemble of several complex CNNs) with the set of parameters $\mathbb{W}_t$ pre-trained for recognizing a high-resolution face $\mc{F}$. Here we assume that $\mathbb{W}_t$ absorbs the rich knowledge encoded in massive high-resolution face images from a teacher face set $\mathbb{D}_t$, in which each face is labeled by an integer from the identity set $\mb{L}_t$. Generally, the number of training face images $|\mb{D}_t|$ is very large and may be invisible to the student stream (\eg, a CNN model released on the Internet is pre-trained with additional face images from private datasets).

\textbf{Student stream}. The student stream $\phi_s(\tilde{\mc{F}};\mathbb{W}_s)$ is a much simpler CNN for recognizing a low-resolution face $\tilde{\mc{F}}$ with parameters $\mathbb{W}_s$. It is learned from the student face set $\mathbb{D}_s=\{(\mc{F}_i,\{\tilde{\mc{F}}_{ij}\}_{j=1}^N,l_i)\}_{i=1}^{|\mathbb{D}_s|}$, where $|\mathbb{D}_s|$ is the number of high-resolution faces. For each high-resolution face $\mc{F}_i$, the student face set also contains its $N$ resolution-degraded versions, and the $j$th resolution-degraded face is denoted as $\tilde{\mc{F}}_{ij}$. Note that both the high-resolution face $\mc{F}_i$ and its degraded versions $\{\tilde{\mc{F}}_{ij}\}_{j=1}^N$ correspond to the same identity label $l_i$ from the identity set $\mb{L}_s$. Here we assume that there are totally $C$ classes of faces in $\mb{L}_s$, and the number of high-resolution faces for the $c$th class is $K_c$ such that $|\mathbb{D}_s|=\sum_{c=1}^{C}{K_c}$.

\subsection{Initialization of the Two-stream CNNs}
As shown in Fig.~\ref{fig:framework}, our two-stream CNNs simultaneously conduct high-resolution and low-resolution face recognition with a teacher stream $\phi_t({\mc{F};\mb{W}_t})$ and a student stream $\phi_s({\tilde{\mc{F}};\mb{W}_s})$, respectively. The parameter set $\mb{W}_t$ of the teacher stream can be initilized by state-of-the-art face recognition models or their ensemble, such as VGGFace~\cite{Parkhi2015BMVC} with VGG16 architecture~\cite{Simonyan2015ICLR}, FaceNet~\cite{Schroff2015CVPR} with GoogLeNet architecture~\cite{szegedy2015cvpr} and VGGFace2~\cite{cao2018fg} with ResNet50 architecture~\cite{He2016CVPR}. As a representative example, we use the architecture of VGGFace~\cite{Parkhi2015BMVC} in the teacher stream and initialize $\mb{W}_t$ with the author-released model. Note that VGGFace is pre-trained on a massive face image dataset $\mb{D}_t$, which we assume is no longer available in the knowledge distillation process.

\begin{figure*}[t]
\begin{center}
   \includegraphics[width=1.0\linewidth]{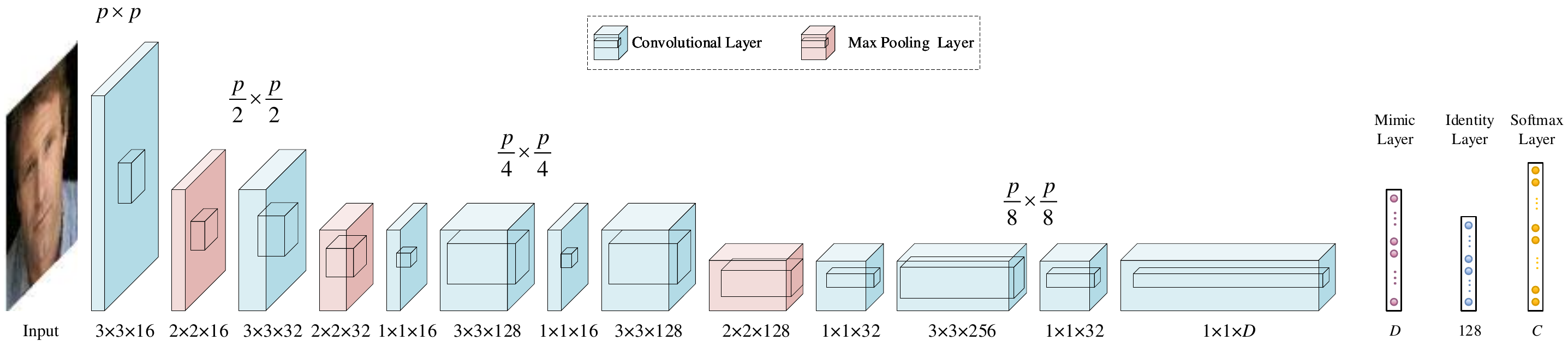}
\end{center}
   \caption{The structure of the proposed student network. It contains ten convolutional layers, three max pooling layers and three fully-connected layers. It only requires $0.79$ million parameters (excluding the last softmax layer), which is much smaller than existing high-resolution face recognition models (\eg, VGGFace). }
\label{fig:network}
\end{figure*}

The student stream $\phi_s(\tilde{\mc{F}};\mathbb{W}_s)$ aims to recognize a low-resolution face $\tilde{\mc{F}}$ with a compact network trained on $\mb{D}_s$. Therefore, we adopt a lightweight network architecture that is similar to \cite{Redmon2017CVPR}. As shown in Fig.~\ref{fig:network}, the student stream can take a low-resolution (\eg,~$32\times32$) face as the input with its majority using $3\times3$ filters and increasing the number of channels after every pooling step. Moreover, the global average pooling is used to make predictions as well as $1\times1$ filters to reduce the feature dimension between $3\times3$ convolutions. Note that a $1\times1\times{}D$ mimic layer is adopted here to receive the knowledge from the teacher stream in the future, where $D$ is the dimension of the learned high-resolution face representation in the teacher stream. Thus, the features from the mimic layer can be used for feature approximation with the teacher network. In addition, since the capacity of the student model is weak, the feature layer that mimics the transferred knowledge should be sufficiently deep, we empirically insert the identity layer between the mimic layer and the softmax layer. The identity layer also plays the role of feature compression.
Finally, the architecture has ten convolutional layers, three max pooling layers and three fully-connected layers. As a result, the amount of parameters in $\mathbb{W}_s$ reaches only $0.79$M, which is only $0.57$\% of the teacher parameter set $\mathbb{W}_s$ (\ie, $138$M). These parameters are first initialized with xavier and then optimized by minimizing the classification loss over $\mb{D}_s$:
\begin{equation}
\begin{split}
\mb{W}_{s}^{*}=\arg\min_{\mb{W}_s}\sum_{i=1}^{|\mb{D}_s|}\sum_{j=1}^{N}{\ell(\phi_s(\tilde{\mc{F}}_{ij};\mb{W}_s),l_i)},
\end{split}
\label{eq:student-init}
\end{equation}
where $\ell(.)$ is a softmax function that measures the classification loss. The minimization problem \eqref{eq:student-init} can be resolved by stochastic gradient descent with standard back-propagation \cite{Krizhevsky2012nips}.

\subsection{Selective Knowledge Distillation from the Teacher Stream}

\begin{figure*}[t]
\begin{center}
   \includegraphics[width=1.0\linewidth]{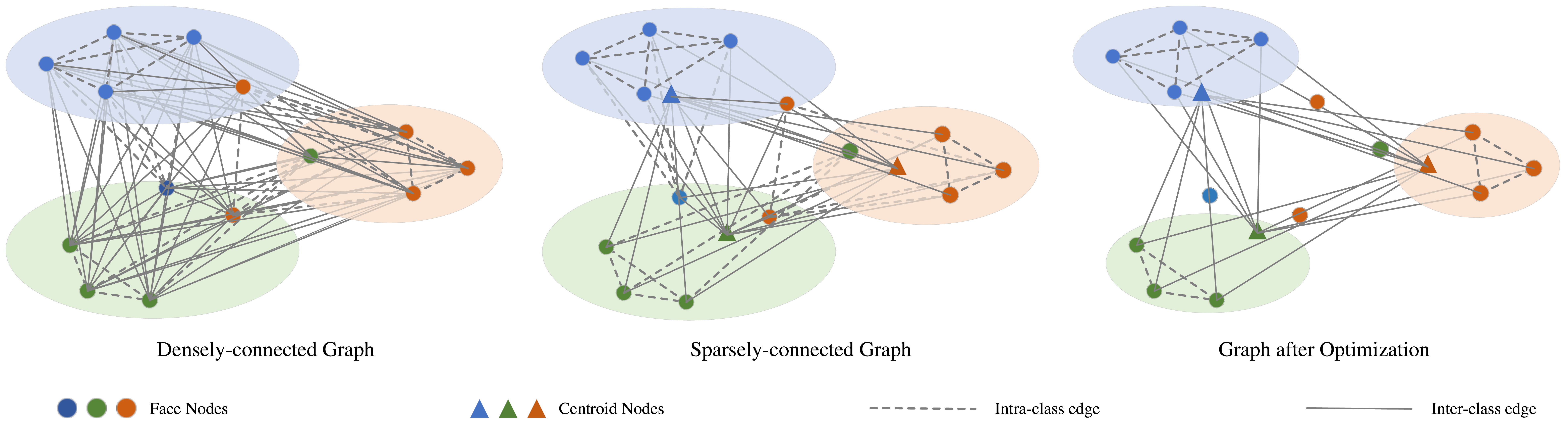}
\end{center}
   \caption{Selective knowledge distillation with graph optimization. The dense-connected graph (left) has too much edges between feature nodes, on which the optimization problem is difficult to be resolved. In order to address that, a sparse-connected graph (middle) is built via adding a virtual centroid node for each node class. Then, the problem is formulated as a graph labeling problem (right) which can be efficiently solved through energy minimization.}
\label{fig:selectivekd}
\end{figure*}

After the initialization stage, the student stream usually suffers from low recognition accuracy over low-resolution faces since many identity cues are missing during the resolution degradation. As a result, its parameters need to be fine-tuned again under the supervision of the teacher stream to learn how to extract the most discriminative features even when the face resolution is very low. Let $\hat{\phi_t}(\mc{F};\hat{\mb{W}}_t)$ and $\hat{\phi_s}(\tilde{\mc{F}};\hat{\mb{W}}_s)$ be the sub-networks composed by the first several layers of the teacher stream $\phi_t(\mc{F};\mb{W}_t)$ and the student stream $\phi_s(\tilde{\mc{F}};\mb{W}_s)$, respectively. $\hat{\phi_t}(\mc{F};\hat{\mb{W}}_t)$ is the feature extraction backend before the softmax layer for extracting the identity features of high-resolution face images, while $\hat{\phi_s}(\tilde{\mc{F}};\hat{\mb{W}}_s)$ corresponds to the main feature branch till the mimic layer and is used to extract approximated features to match the teacher.
The fine-tuning process of the student stream can be described as mimicking the feature representation $\hat{\phi_t}(\mc{F};\hat{\mb{W}}_t)$ with $\hat{\phi_s}(\tilde{\mc{F}};\hat{\mb{W}}_s)$ to improve the final recognition ability of $\phi_s(\tilde{\mc{F}};\mb{W}_s)$. The problem is: how to conduct the inter-network supervision?

Typically, the teacher stream has very powerful ability to recognize high-resolution faces with the identities available in the teacher face set $\mb{D}_t$. However, it can not directly recognize the unfamiliar faces from the student face set $\mb{D}_s$ due to the diverse identities from $\mb{D}_t$ and $\mb{D}_s$. In this case, the knowledge from the teacher stream may contain some errors, which will mislead the fine-tuning process of the student stream. Thus, we selectively distil only the most informative knowledge and reject the wrong one from $\hat{\phi}_t(\mc{F};\hat{\mb{W}}_t)$ to improve the feature extraction ability of the sub-network $\hat{\phi_s}(\tilde{\mc{F}};\hat{\mb{W}}_s)$ and the face recognition ability of the student stream $\phi_s(\tilde{\mc{F}};\mb{W}_s)$.


Toward this end, a feasible solution is finding out the most informative faces from $\mb{D}_s$ by using the features given by the teacher stream, and such informative faces can be defined as the ones with small inter-class similarity and large intra-class similarity. Toward this end, we formulate the selective knowledge distillation process as an inference problem on a graph. The nodes represent faces and the edges represent their correlations. As shown in Fig.~\ref{fig:selectivekd}, a densely-connected graph will contain massive edges between all the nodes from $C$ face classes and thus slow down the inference process. In order to conduct the graph-based inference efficiently, we add a centroid node for each face class and then construct a sparse-connected graph $\mc{G}=\{\mb{V},\mb{E}\}$. In the graph $\mc{G}$, the node set $\mb{V}$ contains two types of nodes: face nodes $\{\mc{F}_i\}_{i=1}^{|\mb{D}_t|}$ and centroid nodes $\{\mc{U}_c\}_{c=1}^{C}$. The $i$th face node $\mc{F}_i$ and the $c$th centroid node $\mc{U}_c$ is represented by $D$-dimensional column feature vectors $\bl{f}_i$ and $\bl{u}_c$, respectively. These two types of feature vectors are extracted from high-resolution face images with the teacher model and can be computed as
\begin{equation}
\bl{f}_i=\hat{\phi}_t(\mc{F}_i;\hat{\mb{W}}_t),~ ~\bl{u}_c=\frac{\sum_{i=1}^{|\mb{D}_s|}\delta(l_i=c)\bl{f}_i}{\sum_{i=1}^{|\mb{D}_s|}\delta(l_i=c)},
\end{equation}
where $\delta(l_i=c)$ is an indicator function which equals 1 if $l_i=c$ and 0 otherwise. We can see that each face node is characterized by the appearance of a specific face, and the centroid node is represented by the average appearance.

With the assistance of centroid nodes, we can construct a sparse graph whose edge set $\mb{E}$ consists of densely-connected intra-class edges $\{(\mc{F}_i,\mc{F}_j)|\forall i,j,l_i=l_j\}$ that link all face nodes within the same class, and sparsely-connected inter-class edges $\{(\mc{F}_i,\mc{U}_c)|\forall i,c,l_i\neq{}c\}$ that link face nodes in one class with the rest centroid nodes outside the class. In this way, the sparsely-connected graph $\mathcal{G}$ contains $(|\mb{D}_s|+C)$ nodes in total and only $\sum_{c=1}^C[{{K_c(K_c-1)}/2+K_c(C-1)}]$ edges.

Given the sparse graph, we can select the most informative faces by solving a binary labeling problem: 
\begin{equation}
\begin{split}
\min_{\bm{\alpha}}&\sum_{i=1}^{|\mb{D}_s|}\left(\alpha_i\sum_{c=1}^{C}\delta(l_i\neq{}c)\cdot{}d(\bl{f}_i,\bl{u}_c)\right)\\
+&\lambda\sum_{i=1}^{|\mb{D}_s|}\sum_{j=1,i<j}^{|\mb{D}_s|}\alpha_i\alpha_j\cdot\delta(l_i=l_j)\cdot{}d(\bl{f}_i,\bl{f}_j),\\
s.t.&~ ~\alpha_i\in\{0,1\}, \forall i=1,\ldots,|\mb{D}_s|,
\end{split}
\label{eq:feature-selection}
\end{equation}
where $\bm{\alpha}=(\alpha_1,\alpha_2,...,\alpha_{|\mb{D}_s|})$ is a binary vector with $|\mb{D}_s|$ components, and its $i$th component $\alpha_i$ equals 1 if the face $\mc{F}_i$ is an informative face and 0 otherwise. Note we use the Cosine distance $d(\cdot)$ to measure the similarity between two feature vectors. We can see that the first term prefers the selection of less informative face nodes that have low similarity with the ``average'' faces in other classes. $\lambda$ is a negative weight that balances the two terms so that the second term prefers the face nodes that have high similarity with other faces with the same identity label. In particular, with the non-negative distance measure $d(\cdot)$ and the negative weight $\lambda$, the first term tends to select less faces and the second term tends to select more. In practice, we can solve the problem \eqref{eq:feature-selection} by using the graph cut algorithm~\cite{Boykov2004PAMI}.

After solving \eqref{eq:feature-selection}, we can select a limited number of informative faces with high intra-class similarity and low inter-class similarity. In this process, the outliers, which are likely to be the errors made by the teacher stream, are discarded from the perspective of feature clustering.
In this way, many helpful knowledge can be accurately distilled and the influences of noisy knowledge introduced by teacher network can be greatly alleviated, which well refines the feature supervision for training the student network. The amount of outliers discarded can be controlled by $\lambda$ (the influence of $\lambda$ will be discussed in experiments).

\subsection{Teacher-supervised Student Stream Fine-tuning}

With the selected informative faces and their features extracted by the teacher stream, the fine-tuning of the student stream will jointly address two issues: 1)~approximating the features of informative faces given by the teacher stream via feature regression, and 2)~recovering the missing facial cues from low-resolution faces. Thus, we can fine-tune the student stream by solving the minimization problem
\begin{equation}
\begin{split}
\min_{\mb{W}_s,\hat{\mb{W}_s}} \mc{L}_{cls}(\mb{W}_s;\mb{D}_s)+\mc{L}_{reg}(\hat{\mb{W}}_s;\mb{D}_s;\bm{\alpha}),
\end{split}
\label{eq:objective}
\end{equation}
where the influences of the classification loss and the regression loss are combined together with equal importance to form a multi-task learning problem. The first term $\mc{L}_{cls}(\mb{W}_s;\mb{D}_s)$ is the classification loss of the student stream over all low-resolution faces. Similar to \eqref{eq:student-init}, it is defined as
\begin{equation}
\begin{split}
&\mc{L}_{cls}(\mb{W}_s;\mb{D}_s)\\
&=\sum_{i=1}^{|\mb{D}_s|}\sum_{j=1}^{N}{\ell(\phi_s(\tilde{\mc{F}}_{ij};\mb{W}_s),l_i)},
\end{split}
\end{equation}
The term $\mc{L}_{reg}(\hat{\mb{W}}_s;\mb{D}_s;\bm{\alpha})$ in \eqref{eq:objective} is the feature regression loss of the sub network $\hat{\phi}_s(\tilde{\mc{F}};\hat{\mb{W}_s})$ formed by feature extraction backend of the student stream. It can be defined as
\begin{equation}
\begin{split}
&\mc{L}_{reg}(\hat{\mb{W}}_s;\mb{D}_s;\bm{\alpha})\\
&=\sum_{i=1}^{|\mb{D}_s|}{\alpha_i}\sum_{j=1}^{N}{\|\hat{\phi}_s(\tilde{\mc{F}}_{ij};\hat{\mb{W}}_s)-\bl{f}_i\|^2}.
\end{split}
\end{equation}
By incorporating these two terms into \eqref{eq:objective}, we can solve the classification and regression tasks via the stochastic gradient descent algorithm with standard back-propagation \cite{Krizhevsky2012nips}. In this way, the student stream can be fine-tuned under the supervision of the teacher stream in the form of feature regression, leading to improved low-resolution face recognition ability with a limited computational cost.

\section{Experiments}

In this section, we first introduce the experiment setting and then conduct four experiments to verify the proposed approach. The first experiment is conducted to analyze and discuss the influence of selective knowledge distillation, and the second experiment compares the performance of teacher and student networks in a face verification task. In the third and the fourth experiments, we further compare the student stream with state-of-the-art low-resolution face models in face recognition task and face retrieval task, respectively. Finally, we conduct the efficiency analysis of the learned student models.

\subsection{Experiment Setting}
We conduct experiments on four well-known face datasets: UMDFaces~\cite{bansal2017umdfaces}, LFW~(Labeled Faces in the Wild)~\cite{Learned2016Labeled}, UCCS~(UnConstrained College Students)~\cite{gunther2017unconstrained} and SCface~(Surveillance Cameras face)~\cite{grgic2011scface}, which are used to verify the proposed approach from the perspective of selective knowledge distillation, face verification, face identification and face retrieval, respectively. Details of the three datasets (and experimental settings) are listed as follows. We implement all the models with TensorFlow~\cite{Abadi2016OSDI} on NVIDIA GPU K80 and single core Intel CPU 2.6G.

The UMDFaces dataset~\cite{bansal2017umdfaces} contains $367,888$ images with annotations from $8,419$ subjects, which is obtained by crawling public images on the Internet. In the experiments, we use this dataset to train all the student models and verify the selective distillation operation. For each training face, we first perform face alignment by using the algorithm in \cite{Ren2014CVPR} to localize facial landmarks. Faces are then cropped and normalized into $224\times{}224$ high-resolution images, which are used as the input of the teacher network. Similarly, to form the low-resolution faces for training each student network, we perform random perturbation $16$ times on the localized facial landmarks, and then crop and normalize the face regions into face images with size $\bm{p}\times{}{\bm{p}}$ where the resolution value $\bm{p}\in\left\{16,32,64,96\right\}$.

On the UMDFaces dataset, all the $224\times{}224$ face images are first fed into VGGFace, the selected teacher stream, to extract $4096$D feature vectors. By solving the graph optimization problem in Eq.~\eqref{eq:feature-selection}, informative features are selected and represented with an indicate vector $\bm{\alpha}$. After that, with the selected informative features, face identity labels and the student input faces, the student network is trained by using standard BP algorithm. In the training, we set the batch size as $256$. Batch normalization layer is introduced to accelerate the network training and prevent over-fitting.

On the LFW dataset~\cite{Learned2016Labeled}, we evaluate all student models in the task of face verification. In the experiment, $6,000$ pairs of face images, including $3,000$ positive and $3,000$ negative pairs, are adopted in the evaluation. The performance is reported as the Area under ROC curve (AUC). In the experiment, feature vectors in the hidden layers (mimic layer and identity layer) are first extracted and normalized from a pair of face images. The similarity between them is calculated for verification by using simple threshold. Unlike~\cite{Sun2014NIPS} that trained Joint Bayesian~\cite{Chen2012ECCV} for face verification, the similarity is used throughout the experiments to directly show the benefit from better supervision utilized to train students.

On the UCCS dataset~\cite{gunther2017unconstrained}, we compare the student models with state-of-the-arts in the face recognition task. Faces from $1,732$ labeled identities subjects are adopted, where blurry, occluded and badly illuminated images are generally common. Note that the identities in training and testing are exclusive. This dataset is suitable to benchmark more challenging unconstrained face recognition in surveillance conditions.

On the SCface dataset~\cite{grgic2011scface}, we compare the student models with state-of-the-arts in face retrieval task. The dataset contains $130$ subjects, each having one high-resolution frontal face image and multiple low-resolution images, captured from three distances (4.2m, 2.6m and 1.0m, respectively) using different quality surveillance cameras. In the experiment, $50$ subjects are randomly selected for training and the rest $80$ subjects for testing. Among the testing images, for each subject, one high-resolution face image is used for constructing retrieval dataset and $15$ low-resolution face images are used for retrieving. As a result, each low-resolution face image from total $80\times{}15=1,200$ is matched with $80$ high-resolution face images. The rank-1 recognition accuracies on three subsets with different distances and total set are reported, respectively.

\subsection{Selective Knowledge Distillation} 
\begin{figure}[t]
  \centering{\includegraphics[width=0.9\linewidth]{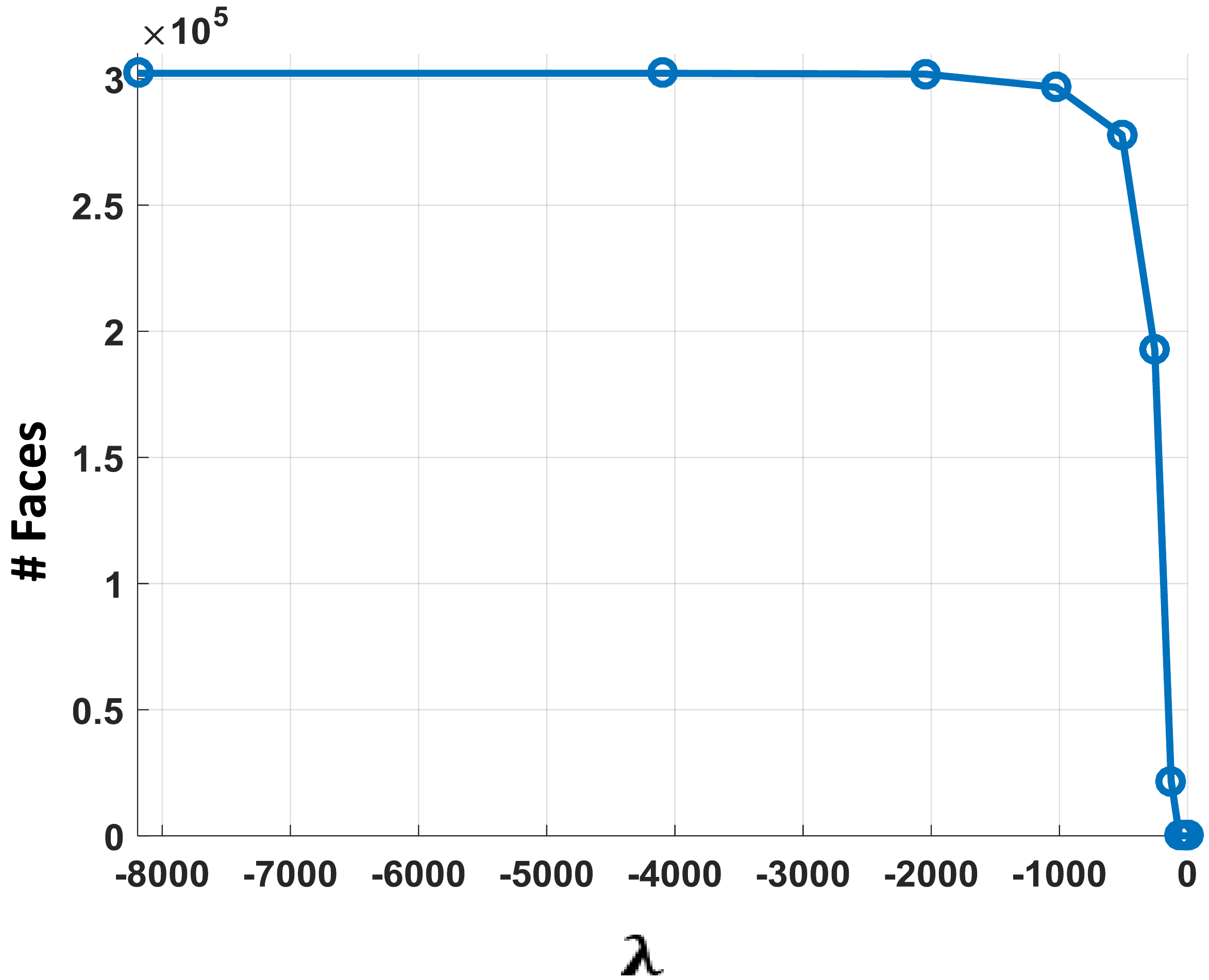}}
  \caption{The influence of the non-positive parameter $\lambda$ to the number of selected informative faces.}
\label{fig:selectivekd-exp}
\end{figure}

To study selective knowledge distillation, first we would like to explore the influence of different settings of parameter $\lambda$ on the parse graph optimization algorithm. Therefore, we gradually increase $\lambda$ from $-8192$ to $0$ with integer power of $2$, and then investigate the decreasing tendency of the number of selected informative faces. In Fig.~\ref{fig:selectivekd-exp}, we show the influence of parameter $\lambda$ on the number of selected informative faces.

When the negative constant $\lambda$ is very small, the number of the informative faces decreases very slowly. After the $\lambda$ increases to around $-1024$, the number of the informative faces starts to decrease sharply. It continues to decrease and remains $0$ after $\lambda$ becomes larger than $-32$.

\begin{figure*}[t]
  \centering{\includegraphics[width=1.0\linewidth]{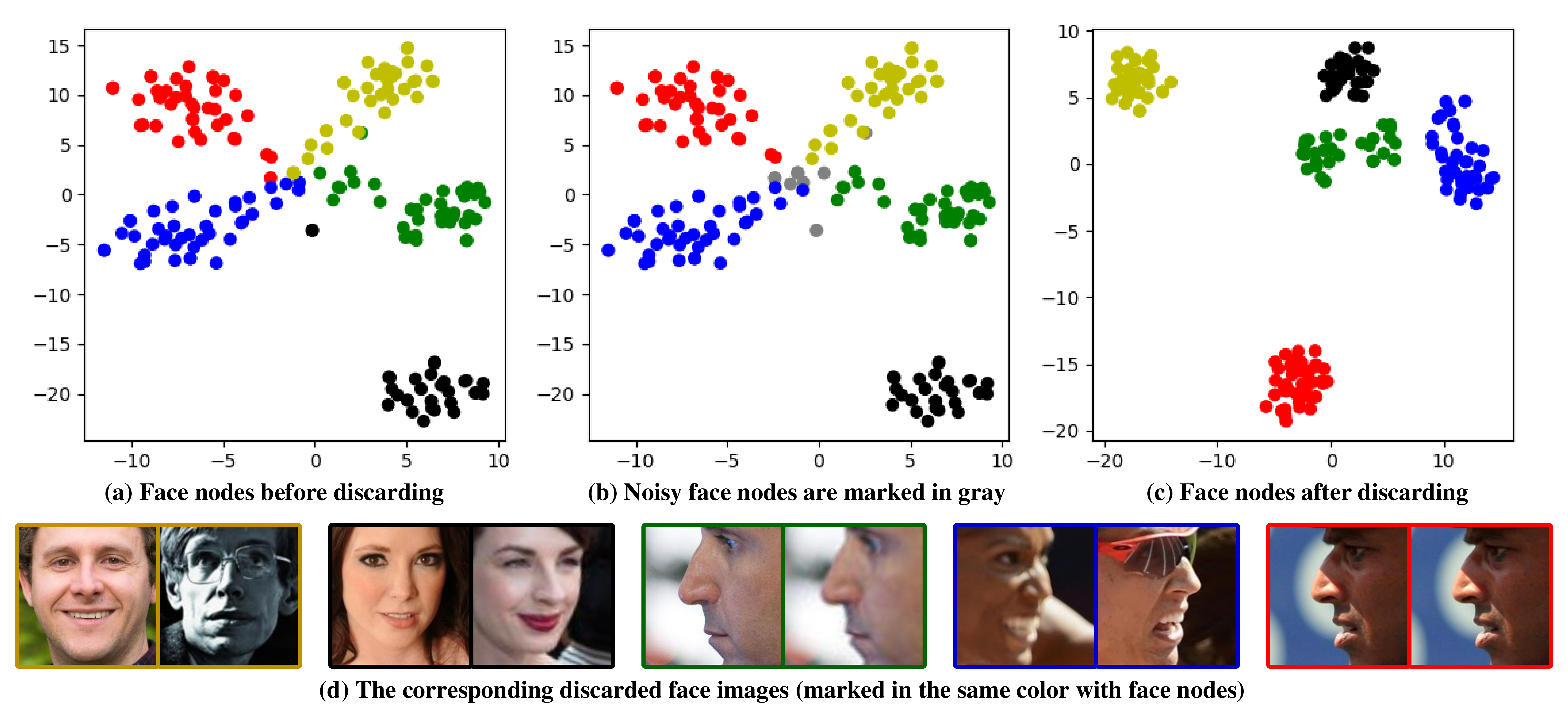}}
  \caption{An example of visualizing selective knowledge distillation on discarding noisy faces with t-SNE~\cite{Maaten2008jmlr}. The noisy face nodes (b) which are far away from their own class or closer to other classes are discarded, which leads to a more compact visualization when performing t-SNE again (c). The corresponding discarded face images (d) usually have side postures, heavy occlusions, inconsistent illumination and blurry appearances, which makes them difficult to be identified by the teacher network.}
\label{fig:selectivekd-exp-class}
\end{figure*}

We further delve into the process of discarding faces during selective knowledge distillation. Fig.~\ref{fig:selectivekd-exp-class} gives an example for showing the process of discarding faces in five identity classes when $\lambda$ increases, where we adopt t-Distributed Stochastic Neighbor Embedding (t-SNE)~\cite{Maaten2008jmlr} to visualize the high-dimensional face nodes. Through solving the sparse graph optimization problem in Eq.~\ref{eq:objective}, some noisy face nodes (see the original face nodes in Fig.~\ref{fig:selectivekd-exp-class}(a) and the noisy nodes in Fig.~\ref{fig:selectivekd-exp-class}(b) marked in gray) that are usually far away from their own class centroid or closer to other classes will be discarded, leading to a more compact visualization effect (see Fig.~\ref{fig:selectivekd-exp-class}(c)). In Fig.~\ref{fig:selectivekd-exp-class}(d), we show the discarded noisy face images, where we can see that the discarded faces are often characterized by side postures, heavy occlusions, inconsistent illuminations or blurry appearances. These challenging images, which may be beyond the recognition capability of the teacher stream, are selected and discarded. This implies that selective knowledge distillation indeed selects the more informative faces while discards the less ones.
Ideally, the teacher model should have a powerful ability to handle various face variations and cluster the faces correctly, which means that high intra-class similarity and low inter-class similarity are achieved with the extracted features by the teacher model. However, in some challenging cases such as large pose variations, the teacher model may fail and thus causes low intra-class similarity (as shown in Fig. \ref{fig:selectivekd-exp-class}). In this case, the extracted teacher knowledge is considered as ``wrong'' and thus will be discarded by our method in feature regression task.

\subsection{Low-Resolution Face Verification on LFW}
With the selected face images, we train many student models to compress the teacher model with different input resolution and various supervision signals. The supervision signals we study are abbreviated as follows:
\begin{enumerate}
\item $c$:  only face class supervision (no distillation).

\item $s$:  selective distillation without face class supervision

\item $sc$: selective distillation with face class supervision

\item $dc$: direct distillation with face class supervision
\end{enumerate}

For the sake of simplicity, a student model is represented as S-$\bm{p}$-$\bm{s}$, where supervision signal $\bm{s}\in\left\{c,s,sc,dc\right\}$. For example, model S-$32$-$sc$ means the student model uses a input resolution of $32\times{}32$ and is trained with both selective distillation and class supervision. Note that S-$\bm{p}$-$c$ is the baseline student model that are directly trained with the supervision of face classes. Similarly, the teacher model is represented as T-$224$ with $224\times{}224$ input. All the student models are trained with the same architecture as shown in Fig.\ref{fig:network}.

\begin{figure}[t]
  \centering{\includegraphics[width=1.0\linewidth]{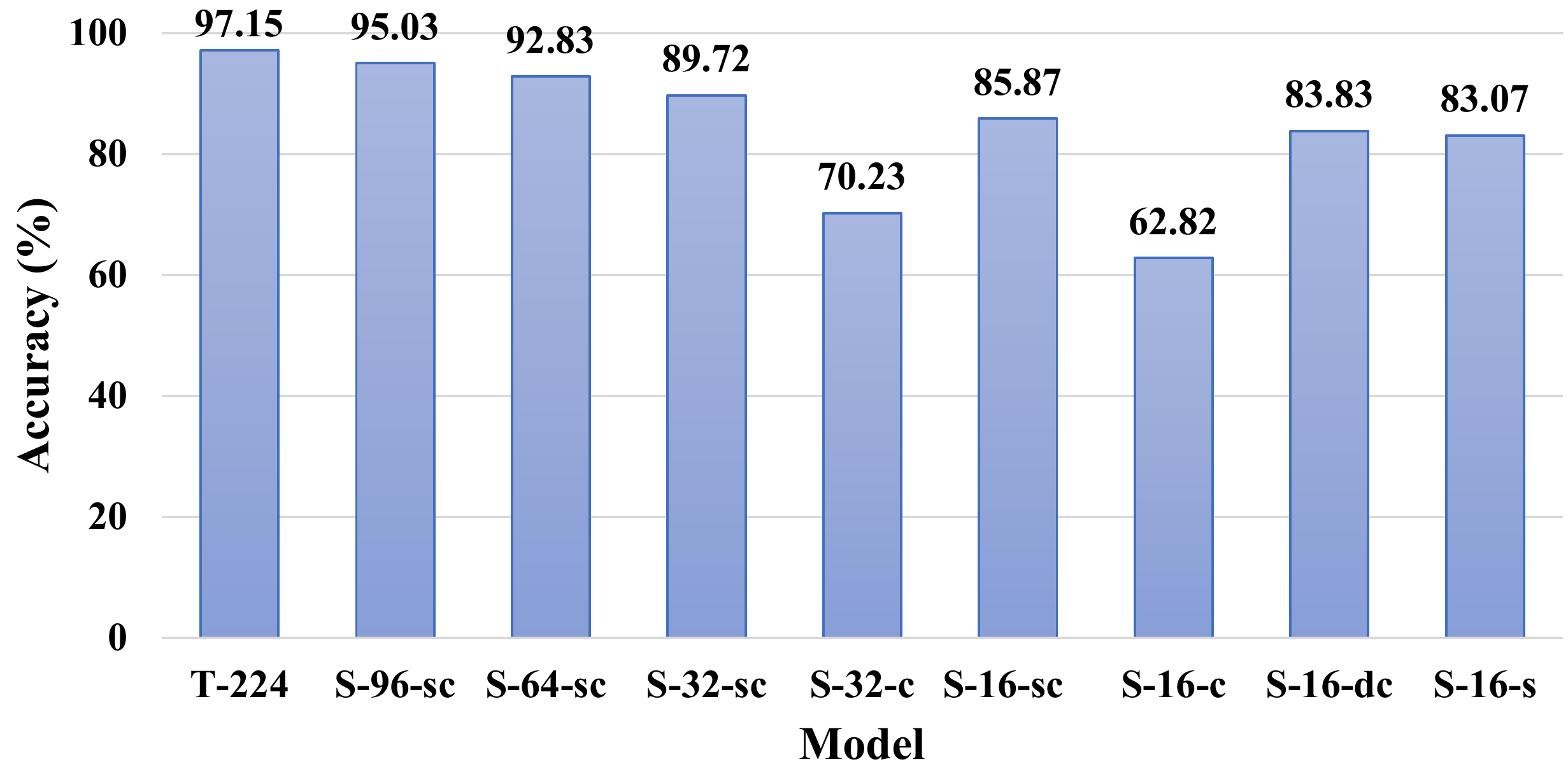}}
  \caption{The recognition accuracy of various teacher and student models on LFW. It shows that the selective knowledge distillation and face class supervision can help to recognizing low-resolution faces.}
  \label{fig:res-lfw}
\end{figure}

The performance of various student models is shown in Fig.~\ref{fig:res-lfw}, from which we can see that the recognition accuracy is decreasing along with the lower face resolution. The student model S-$96$-$sc$ achieves an accuracy of $95.03\%$ by using both selective knowledge distillation and face class supervision, which is only $2.12\%$ lower than the teacher model T-$224$ without metric learning. Note that the model parameter in S-$96$-$sc$ is much less than the teacher model VGGFace ($0.79$M vs. $138$M), and the dimension of face feature vectors has a remarkable compression rate of $32\times$. From these results, we can safely claim that this performance is very competitive particularly for practical deployment on resource-limited devices.

From Fig.~\ref{fig:res-lfw}, we can also find that, without the supervision from the teacher stream, the baseline student model S-$32$-$c$ has a very low accuracy of $70.23\%$, implying that the baseline model itself may lack the capability of extracting discriminative features when being directly trained on low-resolution faces. After being trained with joint supervision signals from the teacher stream and face identities, the model S-$32$-$sc$ achieves a sharp improvement of $19.49\%$ in terms of recognition accuracy (\ie, from $70.23\%$ to $89.72\%$). Similar accuracy improvements can be found between S-$16$-$sc$ and S-$16$-$c$ as well as S-$16$-$dc$ and S-$16$-$c$, implying that either selective or direct knowledge distillation can effectively transfer the teacher's knowledge into the student network so as to improve the recognition performance remarkably.

To further verify the importance of knowledge selection, we compare the performance between S-$16$-$sc$ and S-$16$-$dc$. By carefully selecting informative knowledge, S-$16$-$sc$ achieves an accuracy gain of $2.04\%$ against S-$16$-$dc$ which does not discard noisy faces during training. In addition, the face class supervision signal can also improve the performance, so that the model S-$16$-$sc$ achieves a higher accuracy than S-$16$-$s$. In summary, our two-stream structure can accurately distil informative knowledge from the teacher stream and recover missing knowledge from the student stream.

\subsection{Low-Resolution Face Identification on UCCS}

Since the performance of low-resolution face verification task is promising, we further study low-resolution face identification task on a challenging benchmark, UCCS~\cite{gunther2017unconstrained}, and compare with the state-of-the-art method proposed in~\cite{Wang2016CVPR}, VLRR~(very low-resolution recognition). In VLRR, the cropped face regions are normalized into $80\times{}80$ as high-resolution faces, which are then down-sampled by a factor $5$ for low-resolution images of $16\times{}16$. The evaluation is performed on a $180$-subject subset by layer-by-layer greedy unsupervised model training. Their model reported the best error rates of $40.97\%$ at top-1 and $22.35\%$ at top-5.

Following the experimental settings of \cite{Wang2016CVPR}, we choose a $180$-subject subset of original-resolution images by ranking the subjects according to the number of images. The cropped face regions are then normalized to $16\times{}16$ to obtain $4,825$ images. Note that this number is a little smaller than those claimed by VLRR (\ie, 4,500 training images and 935 testing images). After that, to achieve fair comparisons, we randomly separate the images according to a ratio of $4:1$ to training and testing sets. Finally, we have $3,918$ images for training and the rest $907$ for testing.

On these data, we first train a student model with the input $16\times{}16$ directly on the training set of UCCS and then test the performance on its testing set. In this case, the model achieves $58.65$\% top-1 error rate and $22.71$\% top-5 error rate, which are worse than VLRR. We suspect that the models, once pre-trained on other datasets, can provide valuable prior knowledge on low-resolution visual recognition problem, as stated in \cite{Cheng2018AAAI}. To verify that, we use the student model S-$16$-$sc$ pre-trained on UMDFaces to fine-tune a new model for face identification on UCCS. First, we fix the parameters before mimicking layer and modify the last softmax layer to $180$-way. Then, we train the feature reduction sub-network with the $3,918$ images. The fine-tuned model reaches $32.75\%$ top-1 error rate and $18.3\%$ top-5 error rate, indicating the correct classification of $610$ out of $907$ testing samples in top-1 results and $741$ in top-5, respectively. This implies that our method can achieve better accuracy than VLRR, which may be caused by the fact that the selective supervision from the teacher stream can help the student network learn the discriminative features even when the face resolution is very low.

\begin{figure}[t]
  \centering{\includegraphics[width=1.0\linewidth]{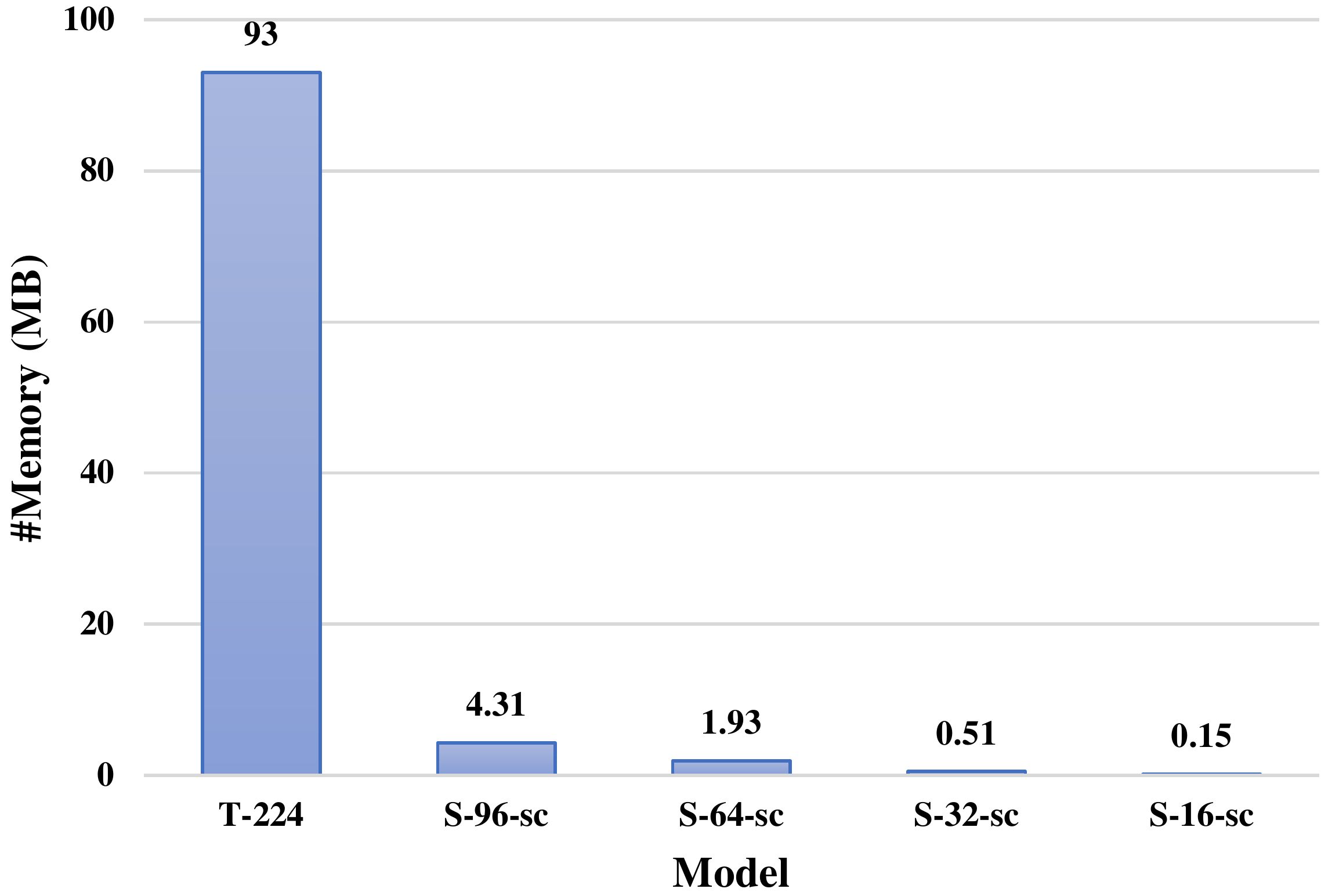}}
  \caption{The inference memory footprint for various models. Compared with the original teacher model, the memory cost of the student model can be greatly reduced.}
  \label{fig:res-param-memory}
\end{figure}

\subsection{Low-Resolution Face Retrieval on SCface}
We further study low-resolution face retrieval task on SCface~\cite{grgic2011scface}, and compare with the baseline (PCA~\cite{grgic2011scface}) and $4$ state-of-the-arts, including three embedding-based models (DCA \cite{haghighat2017low}, DAlign \cite{mudunuri2018dictionary} and LRFRW \cite{li2018low}) and one hallucination-based model (SHSR \cite{singh2018identity}). Here, LRFRW employs deep learning to perform cross-domain transfer. The results are shown in Tab.~\ref{Table:retrieval-accuracy}, where the accuracies on three subsets with different distances and total set are reported respectively. In the experiment, we fine-tune two student models on SCface training set, including S-$64$-$sc$ with the default $128$D identity features and S-$64$-$sc$-$1024$ with $1024$D identity features, respectively.

From the results, we can see several important observations. First, all the models give the accuracies of less than $50\%$ and specially an extremely low total accuracy of $4.73\%$ with the baseline PCA model, showing face retrieval on SCface is a very challenging task. Second, as the resolution increases along with the distance getting closer, the recognition
accuracy gradually increases, which is as expected, which implies that the resolution is indeed an important effect on recognition performance. Thus, the hallucination-based SHSR super-resolves low-resolution face images for feeding to a pre-trained face recognizor, which improves the total accuracy to $16.50\%$. Third, the embedding models which transfer knowledge between different domains, such as transfer features from high-resolution to low-resolution faces by discriminant correlation analysis in DCA model and by supervised discriminative cross-resolution learning in LRFRW model, transfer knowledge from near-infrared to visible images by dictionary alignment in DAlign model, achieve the improved total accuracies of $18.72\%$, $24.30\%$ and $41.04\%$, respectively. This reveals the impact of the transferred knowledge from other domains. Finally, our two models give better total accuracies than other models, \eg, S-64-$sc$ achieves an improved total accuracy of $3.21\%$ over DAlign, implying that the selective knowledge from the teacher stream can facilitate the student network.

\begin{table}[t]
\small
\caption{Rank-1 recognition accuracy (\%) on SCface. Bold and underline indicate the first and second places, respectively.}
\centering{
\begin{tabular}{cccccc}
\toprule
Model                                &~Dist 1~       &~Dist 2~        &~Dist 3~           &~Total~ \\
\midrule
PCA~\cite{grgic2011scface}           &~1.82	         &~6.18	          &~6.18	          &~4.73  \\
SHSR~\cite{li2018low}                &14.70	         &15.70	          &19.10	          &16.50 \\
DCA~\cite{haghighat2017low}          &12.19	         &18.44	          &25.53	          &18.72 \\
LRFRW~\cite{singh2018identity}       &20.40	         &20.80	          &31.71	          &24.30 \\
DAlign~\cite{mudunuri2018dictionary} &34.37	         &39.38	          &\ul{49.37}	  &41.04 \\
S-64-$sc$                            &\ul{39.25} &\ul{45.75}  &47.75	          &\ul{44.25} \\
S-64-$sc$-1024                       &\bl{43.50} &\bl{48.00}  &\bl{53.50}	  &\bl{48.33} \\
\bottomrule
\end{tabular}}
\label{Table:retrieval-accuracy}
\end{table}

\begin{table}[t]
\small
\caption{Inference time and speed on GPU and CPU}
\centering{
\begin{tabular}{ccc}
\toprule
\multirow{2}*{Model}       &~GPU (Nvidia K80)~            &~CPU (Intel 2.6GHZ)~   \\
                           &time / \#faces per second     &time / \#faces per second  \\
\midrule
T-224                      &$~20.4$ ms $ / ~~~49$         &$~982.4$ ms $/ 1.02$  \\
S-96-$sc$                  &$~0.61$ ms $ / 1,639$         &$~~67.9$ ms $/ 14.7$  \\
S-64-$sc$                  &$~0.32$ ms~$ / 3,125$         &$~31.92$ ms $/ 31.3$  \\
S-32-$sc$                  &$~0.15$ ms~$ / 6,667$         &$~~8.92$ ms $/ 112$    \\
S-16-$sc$                  &$0.106$ ms~$ / 9,433$         &$~~2.39$ ms $/ 418$   \\
\bottomrule
\end{tabular}}
\label{Table:inference-time}
\end{table}

\subsection{Efficiency Analysis}
Our approach can greatly reduce the amount of model parameters and memory footprint without significant accuracy drop. As shown in Fig.~\ref{fig:res-param-memory}, the memory reductions are $22\times$, $48\times$, $182\times$ and $620\times$ for the low-resolution student models with $96\times{}96$, $64\times{}64$, $32\times{}32$ and $16\times{}16$, respectively. In particular, for the faces with a very low-resolution of $16\times{}16$, the inference memory is only $0.15$MB.

In addition, the teacher model, VGGFace (T-224), contains $138$ million parameters, while the student network only has about $0.79$ million parameters, making a great reduction of $175\times$ in model complexity at the cost of a very small drop in recognition accuracy. Due to the extreme reductions on the memory cost and the parameter number, the computation complexity can greatly decrease. As shown in Tab.~\ref{Table:inference-time}, the inference runtime on both high-end GPU and low-end CPU is reduced greatly. With a NVIDIA K80 GPU, the inference time for a face is reduced from $20.4$ms with T-224 (VGGFace teacher) to $0.61$ms, $0.32$ms, $0.15$ms and $0.106$ms with S-96-sc, S-64-sc, S-32-sc and S-16-sc, respectively. The inference times are also remarkably reduced even in CPU. Our model takes $0.106$ms and $2.39$ms to recognize a face with a very low-resolution of $16\times{}16$ in GPU and CPU respectively, which means $9,433$ faces per second and $418$ faces per second.

\section{Conclusion}

At present, the problems of large model parameters and high feature dimension widely exist in face recognition models based on deep learning, which hinders their practical deployment on resource-restricted applications (\eg, on embedded or mobile devices). To address this problem, this paper proposes a knowledge distillation method, adopting original large model as the teacher network and letting the teacher selectively supervise the training of student networks via designing the multi-task loss function combining regression and classification items. We have accomplished combination of high-dimensional deep feature regression and low-resolution facial classification, which achieves the uniform compression of deep network and feature dimension with recognition accuracy rate assured. Experimental results show that the proposed approach can transfer the informative knowledge from the teacher network to student models, leading to compact face recognition models with impressive effectiveness and efficiency.

In our future work, we will tentatively explore the usage of recurrent mechanism that aims to handle the failure cases in the teacher stream. Face attributes such as gender, age and makeup will be incorporated into the multi-task framework to further enhance the performance of the compressed model.


\myPara{Acknowledgement}. This work was partially supported by grants from National Key Research and Development Plan (2016YFC0801005), National Natural Science Foundation of China (61772513 \& 61672072), Beijing Nova Program (Z181100006218063), and the International Cooperation Project of Institute of Information Engineering at Chinese Academy of Sciences (Y7Z0511101). Shiming Ge is also supported by Youth Innovation Promotion Association, CAS.

\bibliographystyle{IEEEtran}
\bibliography{bibFD}

%

\begin{IEEEbiography}[{\includegraphics[width=1in,height=1.25in,clip,keepaspectratio]{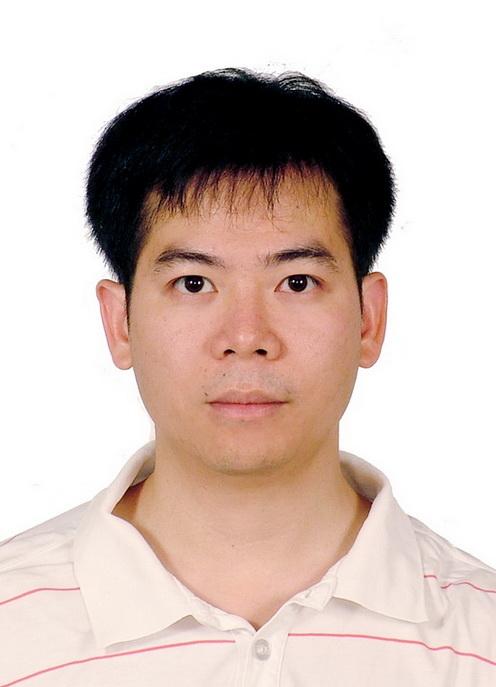}}]{Shiming Ge}
(M'13-SM'15) is currently an Associate Professor at Institute of Information Engineering at Chinese Academy of Sciences. Prior to that, he was a senior researcher in ShanDa Innovations, a researcher in Samsung Electronics and Nokia Research Center. He received the B.S. and Ph.D degrees both in Electronic Engineering from the University of Science and Technology of China (USTC) in 2003 and 2008, respectively. His research mainly focuses on computer vision, deep learning and AI security, especially high-performance deep models towards scalable applications.
\end{IEEEbiography}

\begin{IEEEbiography}[{\includegraphics[width=1in,height=1.25in,clip,keepaspectratio]{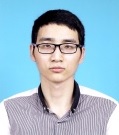}}]{Shengwei Zhao}
received his B.S. degree from the School of Mathematics and Statistics in Wuhan University in 2017. He is now a Master student at the Institute of Information Engineering at Chinese Academy of Sciences and the School of Cyber Security at the University of Chinese Academy of Sciences. His major research interests are deep learning and computer vision.
\end{IEEEbiography}

\begin{IEEEbiography}[{\includegraphics[width=1in,height=1.25in,clip,keepaspectratio]{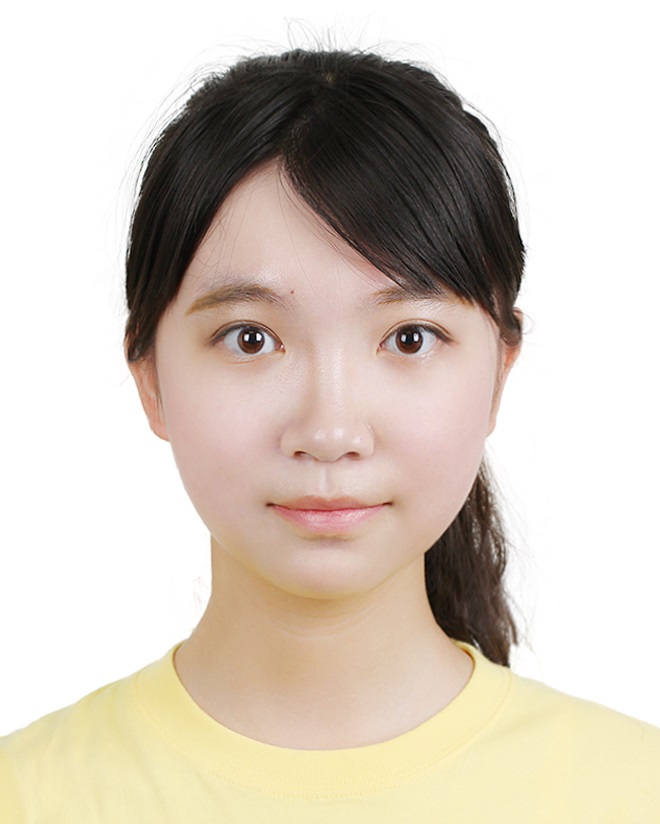}}]{Chenyu Li} is currently a PhD. candidate at the Institute of Information Engineering at Chinese Academy of Sciences and the School of Cyber Security at the University of Chinese Academy of Sciences. She received the B.S. degree from the School of Electronics and Information Engineering at the Tongji University. Her research interests are computer vision and deep learning.
\end{IEEEbiography}

\begin{IEEEbiography}[{\includegraphics[width=1in,height=1.25in,clip,keepaspectratio]{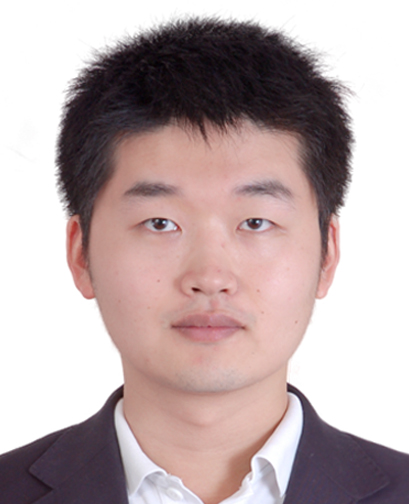}}]{Jia Li} (M'12-SM'15)
is currently an associate Professor with the School of Computer Science and Engineering, Beihang University, Beijing, China. He received the B.E. degree from Tsinghua University in Jul. 2005 and the Ph.D. degree from the Institute of Computing Technology, Chinese Academy of Sciences, in Jan. 2011. Before he joined Beihang University, he used to serve in Nanyang Technological University, Peking University and Shanda Innovations. His research interests include computer vision and multimedia big data, especially the deep learning-based visual content understanding. He is the author or coauthor of over 50 technical articles in refereed journals and conferences such as TPAMI, TIP, IJCV, ICCV and CVPR. His major research interests are cognitive vision towards evolvable algorithms and models.
\end{IEEEbiography}




%
\end{document}